\documentclass[11pt,letterpaper]{article}
\usepackage{naaclhlt2016}
\usepackage{times}
\usepackage{latexsym}
\usepackage{multirow}
\usepackage{graphicx}
\usepackage{caption}
\usepackage{subcaption}
\usepackage{ amssymb }
\usepackage{array}
\usepackage{url}
\usepackage{amsmath}
\usepackage{lipsum}
\usepackage{todonotes}
\usepackage{arydshln}
\usepackage{footnote}
\usepackage{ifthen}
\usepackage{soul}
\usepackage{bm}
\usepackage{flushend}

\naaclfinalcopy % Uncomment this line for the final submission
\def\naaclpaperid{***} %  Enter the naacl Paper ID here

\newcommand{\Note}[2]{} 
\newcommand{\SideNote}[2]{} 
\newcommand{\NoteMRG}[1]{\Note{orange!40}{#1 --MRG}}   
\newcommand{\NoteMY}[1]{\Note{green!40}{#1 --MY}}   
\newcommand{\SideNoteMY}[1]{\SideNote{green!40}{#1 --MY}}

\newcommand{\SideNoteRA}[1]{\SideNote{blue!40}{#1 --RA}}   
\newcommand{\SideNoteMRG}[1]{\SideNote{orange!40}{#1 --MRG}}

\newcommand{\tabref}[1]{Table~\ref{#1}}
\newcommand{\fct}{{\sc fcm}}
\newcommand{\fcm}{{\sc fcm}}
\newcommand{\lrfcm}{{\sc lrfr}}
\newcommand{\lrfcmt}{{\sc lrfr-tucker}}
\newcommand{\lrfcmcluster}{{\sc lrfr$_1$-Brown}}
\newcommand{\lrfcmcp}{{\sc lrfr-cp}}

\newcommand{\lrfcmngram}{{\sc lrfr$_n$}}
\newcommand{\lrfcmtngram}{{\sc lrfr$_n$-tucker}}
\newcommand{\lrfcmcpngram}{{\sc lrfr$_n$-cp}}

\newcommand{\bx}{\mathbf{x}}
\newcommand{\be}{\mathbf{e}}
\newcommand{\bh}{\mathbf{h}}
\newcommand{\bW}[1]{\textrm{W}_{\mkern-4mu #1}}
\newcommand{\bw}{\mathbf{w}}

\newcommand{\cT}{\mathcal{T}}

\newcommand{\y}{\textrm{y}}
\renewcommand{\v}{\textrm{v}}
\newcommand{\w}{\textrm{w}}
\newcommand{\x}{\textrm{u}}

\newcolumntype{H}{>{\setbox0=\hbox\bgroup}c<{\egroup}@{}}

\newcommand{\feat}[1]{\texttt{{\scriptsize #1}}}

\newboolean{FinalVersionWithoutNotes}
\setboolean{FinalVersionWithoutNotes}{false}

\ifthenelse{\boolean{FinalVersionWithoutNotes}}{
\usepackage{comment}
\newcommand{\note}[1]{}

}{

\newcommand{\note}[1]{\hl{[#1]}}

}

\newcommand{\removed}[1]{}
\newcommand{\R}{\mathbb{R}}

\newcommand{\minimize}[3]{
\begin{aligned}
& \underset{#1}{\textrm{minimize:}}
& & #2 \\
& \textrm{subject to:}
& &  #3
\end{aligned}
}

\title{Embedding Lexical Features via Low-Rank Tensors}

 \author{Mo Yu\thanks{Paper submitted during Mo Yu's PhD study at HIT.
 } \\ Harbin Institute of Technology \\  IBM Watson \\ \texttt{yum@us.ibm.com} \And
 	Mark Dredze \\HLTCOE \\ Johns Hopkins University \\ \texttt{mdredze@cs.jhu.edu}
         \AND
         Raman Arora \\ Johns Hopkins University \\ \texttt{arora@cs.jhu.edu} \And
         Matthew R. Gormley \\  Carnegie Mellon University \\\texttt{mgormley@cs.cmu.edu}}
\date{}

\begin{document}

\maketitle

\begin{abstract}
Modern NLP models rely heavily on engineered features, which often combine word and contextual information into 
complex lexical features.
Such combination results in large numbers of features, which
can lead to over-fitting.
% even with the help of recent efforts on usages of word embeddings.
%Word embeddings provide a smoothed representation of words alone.
%When they are combined with non-lexical properties, the large number of features
%can still lead to over-fitting. 
We present a new model that represents complex lexical features---comprised of parts for words, contextual information and labels---in a tensor that captures 
%low dimensional
%representations of each part.
conjunction information among these parts.
We apply low-rank tensor approximations to the corresponding parameter tensors to reduce the parameter space
and improve prediction speed.
Furthermore, we investigate two methods for handling features that include $n$-grams of mixed lengths.
%\SideNoteMY{lengths?}
Our model achieves state-of-the-art results on tasks in relation
extraction, PP-attachment, and preposition disambiguation. 
\SideNoteMRG{The term \emph{parts} here might not be obvious to readers.}
\SideNoteMY{Do you mean that in line 022? If so, we can change to ``comprised of words, contextual information and labels---in a tensor that captures conjunction information among these information.''}
\end{abstract}

\section{Introduction}
\label{sec:intro}

%lexical features are hand designed
%to avoid over-fitting we use embeddings
%the same is true for features. we want to avoid over-fitting those
%previous work did feature embeddings, including NAACL
%we propose a new way of coming up with embeddings by using CP decomposition
%“we use a tucker form approximation on every view”
%multiple lexical parts (bigram) is new
%we show improvements on three different tasks
%this is good because it gives state of the art, and its much faster going from cubic in rank and exponential in the number of lexical parts, to a form that is linear in both rank and lexical parts
%
%include NAACL paper is anonymous supplementary material

\begin{figure*}[htbp]
\centering
\includegraphics[scale=0.47]{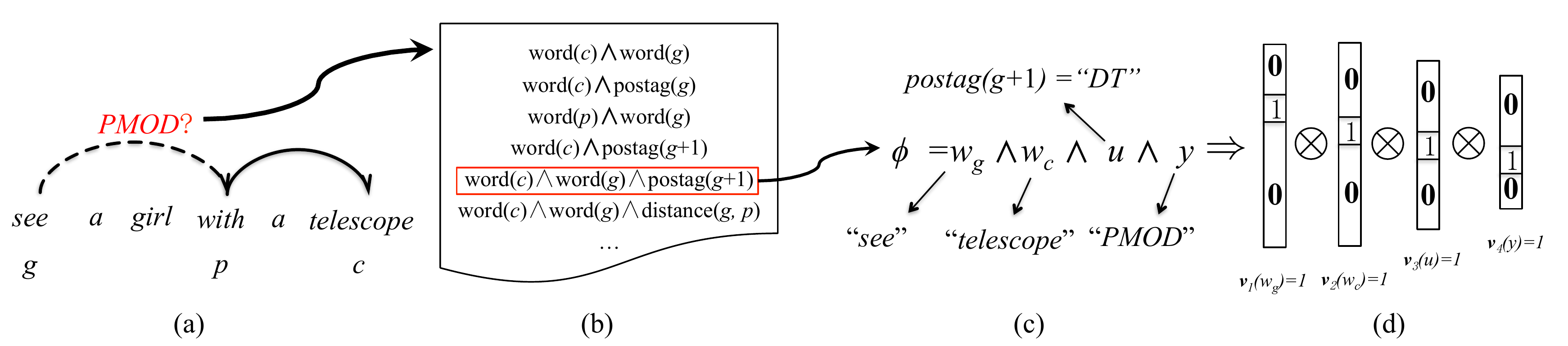}%{featureset.pdf}
%\vspace{-0.05in}
\caption{\small{An example of lexical features used in dependency parsing. To predict the ``PMOD'' arc (the dashed one) between ``see'' and ``with'' in (a),  we may rely on lexical features in (b). Here $p$, $c$, $g$ are indices of the word ``with'', its child (``telescope'') and a candidate head. Figure (c) shows what the fifth feature ($\phi$) is like, when the candidate is ``see''. As is common in multi-class classification tasks, each template generates a different feature for each label $\y$. Thus a feature $\phi=\w_g \wedge \w_c \wedge \x \wedge \y$ is the conjunction of the four parts. Figure (d) is the one-hot representation of $\phi$, which is equivalent to the outer product (i.e. a 4-way tensor) among the four one-hot vectors. $\mathbf{v}(x)=1$ means the vector $\mathbf{v}$ has a single non-zero element in the $x$ position. 
}}
\label{fig:feat_template}
%\vspace{-0.15in}
\end{figure*}

% Statistical models based on supervised learning have become a dominant way on many NLP tasks. 
Statistical NLP models %trained with supervised learning
% Supervised learning techniques are the workhorse for the statistical models that achieve state of the art performance on various NLP tasks. 
% These models rely heavily on feature engineering.
usually rely on hand-designed features, customized for each
task. These features typically combine lexical and contextual information with the label to be scored.
In relation extraction, for example, 
there is a parameter for the presence of a specific relation occurring with a feature conjoining
a word type (lexical) with dependency path information (contextual).
In measuring phrase semantic similarity, a word type is conjoined with its position in the phrase to
signal its role. 
Figure~\ref{fig:feat_template}b shows an example in dependency parsing, where multiple types (words) are conjoined with POS tags or distance information.

%These features, which combine words and other linguistic information, are the hallmark of many tasks.

To avoid model over-fitting that often results from features with lexical components, several smoothed lexical representations have been proposed and shown to improve performance on various NLP tasks; for instance, word embeddings~\cite{bengio2006neural} help improve NER, dependency parsing and semantic role labeling~\cite{miller_name_2004,koo_simple_2008,turian2010word,sun_semi-supervised_2011,roth_composition_2014,hermann-EtAl:2014:P14-1}.
%%Despite widespread use, features with lexical components can lead to over-fitting.
%%causing brittle models that break on new data.
%To avoid model over-fitting that often results from features with lexical components, smoothed lexical representations like
%word embeddings \cite{bengio2006neural}
%%including dimensionality reduction techniques (e.g. PCA), and back-off smoothing of features. 
%%such as word embeddings trained by neural language models 
%% have become a popular tool for learning word representations 
%%as continuous features to mitigate problems
%%of generalization caused by discrete features.
%have improved numerous tasks,
%such as NER, dependency parsing and semantic role labeling
%\cite{miller_name_2004,koo_simple_2008,turian2010word,sun_semi-supervised_2011,roth_composition_2014,hermann-EtAl:2014:P14-1}.
%%,Mo-Yu:2014qv}.
%%Several work also propose to directly learn embeddings for lexical features \cite{ando2005framework,suzuki2009empirical,yang2014unsupervised}.

%By Mo: new motivation paragraph
However, using only word embeddings is not sufficient to represent complex lexical features 
(e.g. $\phi$ in Figure \ref{fig:feat_template}c). In these features, the same word embedding conjoined with 
different non-lexical properties may result in features indicating different labels; the 
corresponding lexical feature representations should take the above interactions into 
consideration. Such important interactions also increase the risk of over-fitting 
% due to exponentially growing feature space, 
as feature space grows exponentially, 
yet how to capture these interactions in representation
learning remains an open question.

%By Mo: I commented the following two paragraphs since they seem not our main motivation

%However, over-fitting is still likely when these embeddings are conjoined with many non-lexical features, 
%hence work on smoothing non-lexical features with embeddings \cite{Yu:2015rt}.

%While word embeddings and feature embeddings are independently useful, complex features require conjunctions 
%of both. \newcite{Mo-Yu:2014qv} proposed a compositional model that combines
%lexical embeddings with arbitrary non-lexical features. \newcite{Yu:2015rt} showed improvements on fine-grained
%relations by adding feature embeddings and forming a single representation.

To address the above problems,\footnote{Our paper only focuses on lexical features, as non-lexical features usually suffer less from over-fitting.} we propose a general and unified approach to reduce the feature space by constructing low-dimensional feature representations, which provides a new way of combining word embeddings, traditional non-lexical properties, and label information.
% learning representations of complex features that include lexical, non-lexical and label
%information. 
Our model exploits the inner structure of features by breaking the feature into multiple parts:
lexical, non-lexical and (optional) label. 
We demonstrate that the full feature is an outer product among these parts. Thus, a parameter tensor
scores each feature to produce a prediction.
Our model then reduces the number of parameters by approximating the parameter tensor with a low-rank tensor: the Tucker approximation of
\newcite{Yu:2015rt} but applied to each embedding type (view), or the 
{\textbf{C}}anonical/{\textbf{P}}arallel-Factors Decomposition (CP).
%to approximate the parameter tensors, which is advantageous
%for features based on multiple lexical parts (e.g. bigrams)
%By Mo: done
%\SideNoteMRG{This is a bit of a run-on, but I didn't edit because of `This' comment below.}
Our models use fewer parameters than previous work that learns 
a separate representation for each feature
\cite{ando2005framework,yang2014unsupervised}.
CP approximation also allows for much faster prediction, going from a method that is
cubic in rank and exponential in the number of lexical parts, to a method linear in both.
Furthermore, we consider two methods for handling features that rely on $n$-grams of mixed lengths.
%Our approach of factorizing a lexical feature and composing its representation
%is general and easy to apply to different tasks.
%Experiments with these models show improvements on three tasks: relation extraction, phrase similarity and PP-attachment.

Our model makes the following contributions when contrasted with prior
work:

\newcite{lei-EtAl:2014:P14-1} applied CP to combine different views of
features. Compared to their work,
our usage of CP-decomposition is different in the application to feature learning: (1) We focus on dimensionality reduction of existing, well-verified features, while \newcite{lei-EtAl:2014:P14-1} generates new features (usually different from ours) by combining some ``atom'' features. Thus their work may ignore some useful features; it relies on binary features as supplementary but our model needs not.
(2) \newcite{lei-EtAl:2014:P14-1}'s factorization relies on views with explicit meanings, e.g. head/modifier/arc in dependency parsing, making it less general. Therefore its applications to tasks like relation extraction are less obvious.

Compared to our previous work \cite{gormley-yu-dredze:2015:EMNLP,Yu:2015rt}, this work allows for higher-order interactions, mixed-length n-gram features, lower-rank representations. We also demonstrate the strength of our new model via applications to new tasks.

%By Mo: replacing the previous two sentences
The resulting method learns smoothed feature representations combining lexical, non-lexical and label information, 
achieving state-of-the-art performance on several tasks: relation extraction, preposition semantics and PP-attachment.

\section{Notation and Definitions}
\label{sec:background}

%\paragraph{Common:}
%The model decompose the structure of $\bx$ to single words.
%For each word $w_i$, a binary vector of features $\mathbf{f_i}$ is defined.
%The binary features $\mathbf{f_i}$ may look at the $i$th word and any other
%substructure of the annotated sentence $\bx$.
%Then if we denote the dense word embedding by
%$e_{w_i}$ and the label-specific model parameters by
%matrix $T_y$,
%e.g. in Fig \ref{fig:example}, the gold label corresponds to matrix $T_y$ where $y$=\textit{ART}$(M_1,M_2)$.

%In the rest of the description, we will take relation extraction as an example. 
%For a pair of entity mentions in a given sentence, the task is to determine the type of relation that holds between the two entities, if any.
%For each pair of mentions in a sentence, we have a training instance $(\bx, y)$; $\bx$ is an annotated sentence, including target entity mentions $M_1$ and $M_2$, and a dependency parse.
%We consider directed relations: for relation type $Rel$,
%$y=Rel(M_1,M_2)$ and $y'=Rel(M_2,M_1)$ are different.
%
%\fcm\ has a log-linear form, which defines a particular
%utilization of the features and embeddings.
%\fcm\ decomposes the structure of $\bx$ into single words.
%For each word $w_i$, a binary feature vector $\mathbf{f_i}$ is defined, which
%considers the $i$th word and any other
%substructure of the annotated sentence $\bx$.
%We denote the dense word embedding by
%$e_{w_i}$ and the label-specific model parameters by
%matrix $T_y$,
%e.g. in Figure \ref{fig:example}, the gold label corresponds to matrix $T_y$ where $y$=\textit{ART}$(M_1,M_2)$.

%\paragraph{Tensor Related: }
We begin with some background on notation and definitions.
%, summarized in \tabref{tab:notations}.
Let $\cT \in \mathbb{R}^{d_1 \times \cdots \times d_K} $ be a $K$-way tensor (i.e., a tensor with $K$ views). In this paper, we consider the tensor $k$-mode product, i.e. multiplying a tensor $\cT \in \mathbb{R}^{d_1 \times \cdots \times d_K}$ by a matrix $\bx \in \mathbb{R}^{d_k \times J}$ (or a vector if $J=1$) in mode (view) $k$. 
The product is denoted by $\cT \times_k \bx$ and is of size $d_1 \times \cdots  \times d_{k-1} \times J \times d_{k+1} \times \cdots \times d_K$. Element-wise, we have 
%\vspace{-0.25in }
%\lipsum*[1]
%\begin{align}
\vspace{-0.15in}
\begin{equation}
(\cT \times_k \bx)_{i_1\ldots i_{k-1}\ j\ i_{k+1}\ldots i_K} = \sum_{i_k = 1}^{d_k} \cT_{i_1 \ldots i_k \ldots i_K} \bx_{i_k j}, \nonumber
\end{equation}
%\end{align}
%\lipsum[1]
%\vspace{-0.15in }
%\SideNoteMY{\raman{We can probably use a better notation for a fiber}}
for $j=1,\ldots,J$. A mode-$k$ fiber $\cT_{i_1\ldots i_{k-1} \bullet  i_{k+1}\ldots i_K}$ of $\cT$ is the $d_k$ dimensional vector obtained by fixing all but the $k$th index. The mode-$k$ unfolding $\cT_{(k)}$ of $\cT$ is the $d_k \times \prod_{i\ne k} d_i$ matrix obtained by concatenating all the $\prod_{i\ne k} d_i$ mode-$k$ fibers along columns. 
%\raman{I think there are $\prod_{i \neq k} n_i$ mode-$k$ fibers.}

Given two matrices $\bW{1} \in \R^{d_1 \times r_1}, \bW{2} \in \R^{d_2 \times r_2}$,  we  write $\bW{1} \otimes \bW{2}$ to denote the Kronecker product between $\bW{1}$ and $\bW{2}$ (outer product for vectors). We  define the Frobenius product (matrix dot product) $\mathbf{A} \odot \mathbf{B}=\sum_{i,j} A_{ij} B_{ij}$ between two matrices with the same sizes; and define
element-wise (Hadamard) multiplication $\mathbf{a} \circ \mathbf{b}$ between vectors with the same sizes.
%\SideNoteMY{\raman{The dimensions are not compatible for the dot product as defined}. 
%For computational reasons, we approximate tensors with following low-rank representations. 
%I think we do not need this definition in this section. And I tried to just define element-wise
%multiplication here}
% For computational reasons, we consider the following representations of tensors. 

\paragraph{Tucker Decomposition:} 
Tucker Decomposition represents a $d_1 \times d_2 \times \ldots \times d_K$ tensor $\cT$ as:
\vspace{-0.03in}
\begin{align} 
{\cT}&= g \times_1 \bW{1} \times_2 \bW{2} \ldots \times_K \bW{K}
\end{align}
where each $\times_i$ is the tensor $i$-mode product and each $\bW{i}$ is a $r_i \times d_i$ matrix. Tensor $g$ with size $r_1 \times r_2 \times \ldots \times r_K $ is called the {\emph{core tensor}. % and $\cT \approx \hat{\cT}$ in the sense that $\sum_{i,j,k} (\cT_{ijk} - \hat{\cT}_{ijk})^2$ is minimized. 
We say that ${\cT}$ has a Tucker rank $(r_{(1)},r_{(2)},\ldots, r_{(K)})$, where $r_{(i)}=\textrm{rank}({{\cT}}_{(i)})$ is the rank of mode-$i$ unfolding. 
To simplify learning, we define the Tucker rank as $r_{(i)}$$=$rank($\mathbf{g}_{(i)}$), which can be bounded simply by the dimensions of $g$, i.e. $r_{(i)} \le r_i$; this allows us to enforce a rank constraint on $\cT$ simply by restricting the dimensions $r_i$ of  $g$, as described in \S\ref{ssec:train}.

% However, to simplify learning, we define the Tucker rank as $r_{(i)}$$=$rank($\mathbf{g}_{(i)}$), which can be bounded simply by the dimensions of $g$, i.e. $r_{(i)} \le r_i$. 
%Then we can remove the rank constrains by
%pre-defining the size $r_i$ of each view $i$ of $g$,
%with a size smaller than $r_i$, 
% A simple way to enforce a rank constraint on  $\cT$ is to predefine the dimensions $r_i$ of  $g$, as described in \S\ref{ssec:train}.
\SideNoteRA{We are not really removing the rank constraint, we should perhaps say that we can express/re-write/simplify the rank constraints. Or simply say ``A simple way to enforce a rank constraint on the tensor $\cT$ is to constrain the dimensions of the core tensor $g$, as described in \S...''.}
\SideNoteMY{fixed}
%\SideNoteMY{ \raman{How does redefining rank simplify the learning ?}} 
% Tucker decomposition yields a low-rank approximation $\cT \approx \hat{\cT}$ of  in the sense that $\sum_{i,j,k} (\cT_{ijk} - \hat{\cT}_{ijk})^2$ is small. 

\paragraph{CP Decomposition:} CP decomposition represents a $d_1$$
\times$$ d_2$$ \times $$\ldots $$\times$$ d_K$ tensor $\cT$ as a sum
of rank-one tensors (i.e. a sum of outer products of $K$ vectors):
\SideNoteMRG{This was oddly phrased with a double colon. Switched to parenthetical.}
\vspace{-0.05in}
\begin{align}
%{\cT}&= \bW{1} \otimes \bW{2} \otimes \ldots \otimes \bW{K}  \label{eq:cp_decomp} \\
{\cT}= \sum_{j=1}^r \bW{1}[j,:] \otimes \bW{2}[j,:] \otimes \ldots \otimes \bW{K}[j,:]\label{eq:cp_decomp}
\end{align}
where % $\cT \approx \hat{\cT}$ as above, 
each $\bW{i}$ is an $r \times d_i$ matrix and $\bW{i}[j,:]$ is the vector of its $j$-th row. For CP decomposition, the rank $r$  of a tensor ${\cT}$ is defined to be the number of rank-one tensors in the decomposition. CP decomposition can be viewed as a special case of Tucker decomposition in which
$r_1=r_2=\ldots=r_K=r$ and
$g$ is a superdiagonal tensor.
% ($g_{i_1i_2\ldots i_K}=1$ only if $i_1=i_2=\ldots=i_K$, otherwise $g_{i_1i_2\ldots i_K} = 0$).
%\section{Three Steps for Embedding Lexical Features}

\NoteMRG{By the end of the Background section, I don't have any
  intuition for why the Tucker decomposition is a good idea. At this point, we
  don't have space to address this. However, the camera-ready should
  probably try to.}
\NoteMY{In my opinion sometimes Tucker will be better since the ranks are controlled by variables $r_i$, so sometimes it will be easier to learn since we always have $r_i < d_i$
But for CP, the good approximation may be achieved at some r, where r is significantly larger than all the $d_i$s}

\section{Factorization of Lexical Features}
\label{lab:factor}

Suppose we have feature $\phi$ that includes information from a label $\y$, multiple lexical items $\w_1,\ldots,\w_n$ and non-lexical property $\x$.
This feature can be factorized as a conjunction of each part: $\phi=\y \wedge \x \wedge \w_1 \wedge \ldots \wedge \w_n$.
%\footnote{
%%This factorization is easy to achieve in a task with a given feature template,
%%as illustrated in
%See Appendix C, D, E for examples of such factorization.}
The feature fires when all $(n+2)$ parts fire in the instance 
(reflected by the $\wedge$ symbol in $\phi$).
The one-hot representation of $\phi$ can then be viewed as a tensor $e_{\phi} = \y \otimes \x \otimes \w_1 \otimes \cdots  \otimes \w_n$, where each feature part is also represented as a one-hot vector.\footnote{$\x, \y, \w_i$ denote one-hot vectors instead of symbols.}
%
%If we represent each feature part as a one-hot vector, then $\phi = \y \otimes \x \otimes \w_1 \otimes \cdots  \otimes \w_n$. 
Figure~\ref{fig:feat_template}d illustrates this case with two lexical parts.

Given an input instance $\bx$ and its associated label $\y$, 
we can extract a set of features $S(\bx, \y)$.
In a traditional log-linear model, we view the instance $\bx$ as a bag-of-features, i.e. a feature vector $F(\bx, \y)$. Each dimension corresponds to a feature $\phi$, and has value 1 if $\phi \in S(\bx, \y)$. Then the log-linear model scores the instance as $s(\bx,\y; \bw) = \bw^T F(\bx, \y) =\sum_{\phi \in S(\bx,\y)} s(\phi;\bw) $,
where $\bw$ is the parameter vector. 
We can re-write $s(\bx,\y; \bw)$ based on the factorization of the features using
tensor multiplication; in which $\bw$ becomes a parameter tensor $\cT$:
%represented as a tensor $\cT$, the scoring function is given as 
% according to the feature factorization: 
%Our basic model scores an instance with a parameter tensor $T$.
%\vspace{-0.1in}
\begin{align}
%P(y|\bx;\cT) = \frac{\exp \left \{ \sum_{\phi \in F(\bx,y)} s(\phi;\cT)  \right \}}{ \sum_{y' \in L} \exp \left \{ \sum_{\phi \in F(\bx,y)} s(\phi;\cT)  \right \}}, \nonumber
%s(\bx, \y; \bw) &= \sum_{\phi \in S(\bx,\y)} s(\phi;\bw) & \nonumber  \\
%= s(\bx,\y;\cT) &= \sum_{\phi \in S(\bx,\y)} s(\phi;\cT) & \label{eq:score_function}
s(\bx, \y; \bw) = s(\bx,\y;\cT) = \sum_{\phi \in S(\bx,\y)} s(\phi;\cT) & \label{eq:score_function}
\end{align}
%\textrm{where } &
Here each $\phi$ has the form $(\y, \x, \w_1, \ldots, \w_n)$, and
%\vspace{-0.1in}
\begin{align}
%s(\phi; \cT) = \cT \times_w w_{\phi} \times_f f_{\phi} \times_l l_{\phi}.
s(\phi; \cT) = \cT \times_l \y \times_f \x \times_{\w_1} \w_1 ... \times_{\w_n} \w_n.
\label{eq:basic}
\end{align}

Note that one-hot vectors $\w_i$ of words themselves are large ($|\w_i|>$ 500k), thus the above
formulation 
%While this formulation allows for factorizing $\phi$, it 
with parameter tensor $\cT$ can be very large, making parameter estimation difficult.
%it is difficult to have sufficient labeled data to estimate the huge number of entries in $\cT$. 
Instead of estimating only the values of the dimensions which appear in training data as in traditional methods,
%Traditional method only evaluate the values of the dimensions which appear in training data.
%Instead, 
we will reduce the size of tensor $\cT$ via a low-rank approximation.
With different approximation methods,  \eqref{eq:basic} will have
different equivalent forms, e.g. \eqref{eq:fcm_ngram},
\eqref{eq:fcmcp_ngram} in \S
\ref{sec:lr_model_tucker}.\SideNoteMRG{The phrase \emph{different
    equivalent} seems odd. Maybe just \emph{different}?}
\SideNoteMY{Is ``equivalent forms'' a usual phrase in maths? I am trying to use ``different'' to modify the term ``equivalent-forms''}
%Our general  dimension reduction framework, i.e. \emph{Low-Rank Feature Representation} (\lrfcm),  is shown below. The methods in Section~\ref{sec:lr_model_tucker} fall into this framework with different low-rank forms of tensors.

%\begin{figure}[t]
%\centering
%\includegraphics[scale=0.4]{feat_factorization.pdf}
%\vspace{-0.35in}
%\caption{\small{Illustration of our feature factorization. 
%%The feature $\phi$ is from Figure \ref{fig:feat_template}c.
%%, which fires when all the four parts fire in the instance 
%%(reflected by the $\wedge$ symbol in $\phi$).
%%$\phi=\w_g \wedge \w_c \wedge \x \wedge \y$ reflects the above conjunction relation.
%The one-hot representation of $\phi$ (from Figure \ref{fig:feat_template}c) is equivalent to the outer product (i.e. a 4-way tensor) among the four one-hot vectors on the right side. $\mathbf{v}(x)=1$ means that the vector $\mathbf{v}$ has a single non-zero element on the $x$ position. 
%%This is equivalent to a 4-way tensor. 
%}}
%\label{fig:feat_repr}
%\vspace{-0.15in}
%\end{figure}

\paragraph{Optimization objective:} 
%Using this factorization of the score function, the overall score for an instance
%s $s(\bx,y;\cT) = \sum_{\phi \in F(\bx,y)} s(\phi;\cT)$.
The loss function $\ell$ for training the log-linear model uses \eqref{eq:score_function} for scores, 
%Based on the scores, we can compute the loss function $\ell$ on each training instance $(\bx,y)$. 
e.g., the log-loss $\ell(\bx, \y; \cT)=-\log \frac{\exp\{s(\bx,\y;\cT)\}}{\sum_{\y' \in L}\exp\{s(\bx,\y';\cT)\}}$.
Learning can be formulated as the following optimization problem: %find $\cT$ that solves
%\vspace{-0.1in}
\begin{align}
%\small
%\minimize{\cT}{\sum_{(\bx,y)\in \mathcal{D}} \ell(y, \arg \min_{y'} {s} (F(\bx,y');\cT))}{\left\{\begin{matrix}
\minimize{\cT}{\sum_{(\bx,y)\in \mathcal{D}} \ell(\bx, y;\cT)}{\left\{\begin{matrix}
\textrm{rank}({\cT}) \le (r_1,r_2,...,r_{n+2}) \\
\text{\quad\quad\quad\quad\quad (Tucker-form)} \\
\textrm{rank}({\cT}) \le r \text{\quad(CP-form)} 
\end{matrix}\right.}
\label{eq:objective}
\end{align}
%
%\begin{align}
%\arg \max_{\hat{T}} \sum_{(\bx,y)\in \mathcal{D}} \ell(y, \arg \min_{y'} \hat{T} (F(\bx,y')))
%\label{eq:objective}
%\end{align}
%\vspace{-0.25in}
%\begin{align}
%s.t.: 
%\left\{\begin{matrix}
%\textrm{rank}(\hat{T}) \le (r_1,r_2,r_3) \text{ (Tucker-form)} \\
%\textrm{rank}(\hat{T}) \le r \text{\quad\quad\quad(CP-form)}
%\end{matrix}\right. \nonumber
%%rank(\hat{T}) \le r &\text{(CP-form)} \nonumber \\
%%&\text{or} rank(\hat{T}) \le (r_1,r_2,r_3) & \text{(Tucker-form)}
%\end{align}
%
%where $\ell$ is the loss function about the prediction of $y$ given $\bx$ with model parameter $\cT$.
where the constraints on rank($\cT$) depend on the chosen tensor approximation method (\S \ref{sec:background}).

The above framework has some advantages:
First, as discussed in \S\ref{sec:intro} and here, 
we hope the representations capture rich interactions between different parts of the lexical features; the low-rank tensor approximation methods keep the most important interaction information of the original tensor, while significantly reducing its size.
Second,
%In addition to learning weights $\cT$ that minimize the loss function $\ell$, 
the low-rank structure will 
encourage weight-sharing among lexical features with similar decomposed parts, leading to 
better model generalization.
Note that there are examples where features have different numbers of multiple lexical parts, such as both unigram and bigram features in PP-attachment.
We will use two different methods to handle these features 
(\S\ref{sec:lr_model_ngram}).
% as evidenced by our empirical results (\S \ref{sec:exp}).
%The effect of the optimization problem above is to give two features with similar lexical, non-lexical parts and labels similar representations. 
%%(i.e. those with similar embeddings) with
%%similar functions in the sentence (i.e. those with similar features)
%%similar matrix representations.
%The low-rank structure encourages weight sharing and thus helps the model generalize better as we show in the empirical results in Section~\ref{sec:exp}. 
% In this way they can share some common model parameters thus the model can better generalize.
%Thus, this model generalizes its model parameters across words with similar embeddings
%only when they share similar functions in the sentence.
%For the example in Figure \ref{fig:example}, we hope to
%learn parameters which give words similar to ``driving'' with the {\tt is-on-dependency-path} feature and  $\wedge$ type($M_1$)=PER $\wedge$ type($M_2$)=VEH} ) high weight for relations like \textit{ART} labels.

%\textcolor{red}{
%In practice, usually we will use feature sets consisting of features with different lexical parts, e.g. we may use both unigram and bigram features in dependency parsing. With the factorization method
%described in this section, these features will be decomposed into different numbers of parts (views).
%To use all these heterogeneous features in a feature set, we propose two extensions of above method in \S\ref{ssec:lr_ngram} and \S\ref{ssec:lr_clusters}.}
%
%
%

\paragraph{Remarks (advantages of our factorization)}
Compared to prior work, e.g. \cite{lei-EtAl:2014:P14-1,lei-EtAl:2015:NAACL-HLT}, the proposed factorization has the following advantages:

\begin{enumerate}
\item{\textbf{Parameter explosion} when mapping a view with lexical
    properties to its representation vector (as will be discussed in
    \ref{ssec:lr_embedding}): Our factorization allows the model to
    treat word embeddings as inputs to the views of lexical parts,
    dramatically reducing the parameters. Prior work cannot do this
    since its views are mixtures of lexical and non-lexical
    properties. Note that \newcite{lei-EtAl:2014:P14-1} uses
    embeddings by \textit{concatenating} them to specific views, which
    increases dimensionality, but the improvement is limited.}

\item{\textbf{No weight-sharing} among conjunctions with same lexical
    property, like the child-word ``word($c$)'' and its conjunction
    with head-postag ``word($c$) $\wedge$ word($g$)'' in
    Figure~\ref{fig:feat_template}(b). The factorization in prior
    work treats them as \textit{independent} features, greatly
    increasing the dimensionality. Our factorization builds
    representations of both features based on the embedding of
    ``word($c$)'', thus utilizing their connections and reducing the
    dimensionality.}
\end{enumerate}
The above advantages are also key to overcome the problems of prior work mentioned at the end of \S\ref{sec:intro}.

\section{Feature Representations via Low-rank Tensor Approximations}

Using one-hot encodings for each of the parts of feature $\phi$ results in a very large
tensor.
This section shows how to compute the score in \eqref{eq:basic} without constructing the full feature tensor
using two tensor approximation methods (\S\ref{sec:lr_model_tucker} and \S\ref{ssec:lr_ngram}).

We begin with some intuition.
To score the original (full rank) tensor representation of $\phi$, we need a parameter tensor $\cT$ of size $d_1 \times d_2 \times \ldots \times d_{n+2}$, where $d_3= \cdots =d_{n+2}=|V|$ is the vocabulary size, % size of word vocabulary, 
$n$ is the number of lexical parts in the feature and $d_1= |L|$ and
$d_2 = |F|$ are the number of different labels and non-lexical
properties, respectively. 
(\S \ref{sec:lr_model_ngram} will handle $n$ varying across features.)
Our methods reduce the tensor size by embedding each part of $\phi$ into a lower dimensional space, where we represent each label, non-lexical property and words with an $r_1, r_2$, $r_3,\ldots, r_{n+2}$ dimensional vector respectively ($r_i \ll d_i$,  $\forall i$). 
% where we represent each word $\w$ with an $r_1$-dimensional vector, each unlexical feature $\x$ with an $r_2$-dimensional vector and label $\y$ with an $r_3$ dimensional vector. 
These embedded features can then be scored by much smaller tensors.
%This results in $r_1 \times \cdots \times r_{n+2}$ tensor $\cT$.
We denote the above transformations as matrices $\bW{\l} \in \R^{r_{1} \times d_{1}}$,  $\bW{f} \in \R^{r_{2} \times d_{2}}$, $\bW{i} \in \R^{r_{i+2} \times d_{i+2}}$ for $i=1,\ldots,n$, and write corresponding low-dimensional hidden representations as $\bh^{(\l)}_\y=\bW{\l} \y$, $\bh^{(f)}_\x = \bW{f} \x$ and $\bh^{(i)}_\w=\bW{i} \w$. 

In our methods, the above transformations of embeddings are \emph{parts of low-rank tensors} as in \eqref{eq:objective}, so the embeddings of non-lexical properties and labels can be trained simultaneously with the low-rank tensors.
Note that for one-hot input encodings the transformation matrices are essentially lookup tables, making the computation of these transformations sufficiently fast.
%We denote the lookup tables for these embeddings with matrices $\bW{\ell}^{r_{1} \times d_{1}}$,  $\bW{f}^{r_{2} \times d_{2}}$ and $\bW{i}^{r_{i+2} \times d_{i+2}}$ and corresponding low-dimensional embeddings as $\bh^{(\ell)}_\y=\bW{\ell}(:,\y)$, $\bh^{(f)}_\x = \bW{f}(:,\x)$ and $\bh^{(i)}_\w=\bW{i}(:,\w)$.
\SideNoteRA{I have changed the notation here as it was not clear earlier with the same symbol indexing the lookup matrix and the column of the matrix. Please confirm that you are okay with this notation and check for consistency in the rest of the paper.} 
% We denote the lookup tables for these embeddings with matrices $\bW{\ell}^{r_{1} \times d_{1}}$,  $\bW{f}^{r_{2} \times d_{2}}$ and $\bW{i}^{r_{i+2} \times d_{i+2}}$ and corresponding low-dimensional embeddings as $\bh_\y=\bW{\ell}(:,l)$.

%We now describe two low-rank tensor approximations of this model.
%In this section we propose to use a Tucker form low-rank tensor to approximate the (possibly) full-rank tensor $T$ in Eq. \ref{eq:basic}.

%\subsection{Embedding Different Views of Complex Lexical Features}
% \subsection{Feature Representations with Tucker Form Tensor Approximation}

\subsection{Tucker Form}
\label{sec:lr_model_tucker}
For our first approximation, we 
assume that tensor $\cT$
has a low-rank Tucker decomposition: $\cT = g \times_{\l} \bW{\l} \times_f \bW{f}  \times_{\w_1} \bW{1}  \times_{\w_2}\cdots \times_{\w_n} \bW{n}$. 
 \removed{
Then suppose we have a small core tensor $g$, such that $\cT$ has a low-rank Tucker decomposition
$\hat{\cT} = g \times_\l \bW{\l} \times_f \bW{f}  \times_{\w_1} \ldots \times_{\w_n} \bW{n}$. }
We can then express the scoring function \eqref{eq:basic} for a feature $\phi=(\y, \x, \w_1, \ldots \w_n)$ with $n$-lexical parts, as:
% The scoring function, for a feature $\phi=(\y, \x, \w_1, \ldots \w_n)$ with $n$-lexical parts, given in equation~(\ref{eq:basic}) can be expressed as:
%
%The scoring function requires a small core tensor which is obtained by solving equation~\ref{eq:objective} for Tucker decomposition 
%In the formula in Eq. 3, if $\cT$ has a low-rank Tucker decomposition $\cT = g \otimes \bW{1} \otimes \bW{2} \otimes \bW{3}$ then the scoring function of $\phi = (\w, \x, \y)$ can equivalently be expressed as:
%
%We can now score the feature with $n$-lexical parts $(w_1, \ldots \w_n,\x,\y)$ using the following tensor product (replacing equation~\ref{eq:basic}):
%\begin{align}
%\!s(\w,\!\x,\!\y; g,\!\! \bW_{\{ e,\x,l \}})\!=\!g\! \times_\ell\! \bh_\y\! \times_f\! \mathbf{h_f}\! \times_\w\! \bh_\w.\!\!\! 
%\label{eq:tucker_basic}
%\end{align}
%
%\begin{eqnarray}
%% \textstyle \!P(\y| \bx;\! \mathbf{e},\! g,\!\mathbf{\{\w_i\}}, \!\mathbf{\w_f}, \!\mathbf{\w_\ell}) \! \propto\! \exp\Big(\sum_i g \times_\ell  \nonumber \\ 
%  \textstyle \!s({\w}_{1,\cdots,n}, \x, \y;\!, g,{\{\bW{i}\}},{\bW{f}},{\bW{\ell}})  = g \times_\ell \nonumber\\ 
% \quad \bh_\y \times_f \mathbf{h}_f \times_{\w_1} \bh_{\w_1} \times_{\w_2} \bh_{\w_2}\cdots\times_{\w_n} \bh_{\w_n}. 
%  \label{eq:fcm_ngram}
%\end{eqnarray}
%
\begin{align}
&\hspace*{-14pt}\textstyle \!s(\y, \x, \w_1,\cdots, \w_n; g,{\bW{\l}},{\bW{f}}, {\{\bW{i}\}_{i=1}^n})  \nonumber\\ 
=&\ g \times_{\l} \bh^{(\l)}_\y \times_f \mathbf{h}^{(f)}_\x  \times_{\w_1} \bh^{(1)}_{\w_1} \cdots\times_{\w_n} \bh^{(n)}_{\w_n},
  \label{eq:fcm_ngram}
\end{align}
%
%The scoring function requires a small core tensor which is obtained by solving equation~\ref{eq:objective} for Tucker decomposition $\hat{\cT} = g \times_\ell \bW{\ell} \times_f \bW{f}  \times_{\w_1} \ldots \times_{\w_n} \bW{n}$. 
%In the formula in Eq. 3, if $\cT$ has a low-rank Tucker decomposition $\cT = g \otimes \bW{1} \otimes \bW{2} \otimes \bW{3}$ then the scoring function of $\phi = (\w, \x, \y)$ can equivalently be expressed as:
%\begin{align}
%& \cT \times_\w \w_{\phi} \times_f f_{\phi} \times_\y y_{\phi} \nonumber \\
%= & \left( g \otimes \bW{1} \otimes \bW{2} \otimes \bW{3}  \right) \times_\w \w_{\phi} \times_f f_{\phi} \times_\y y_{\phi} \nonumber \\
%=& g \times_\w (\bW{e} \w_{\phi})  \times_f (\bW{f} f_{\phi})  \times_\y (\bW{\y} y_{\phi}) .
%\end{align}
which amounts to first projecting $\x, \y$, and $\w_{i}$ (for all $i$) to lower dimensional vectors 
$\bh^{(f)}_{\x}, \bh^{(\l)}_{\y}, \bh^{(i)}_{\w_i}$, 
%$\bh^{(i)}_{\w_i} = \bW{i} \w_{i}, \bh^{(f)}_\x = \bW{f} \x, \bh^{(\l)}_\y = \bW{\l} \y$, 
and then weighting these hidden representations using the flattened core tensor $g$. 
% taking all projected hidden representations $\bh$s features and weighting each of these $\bh$s by the flattened core tensor $g$. 
The low-dimensional representations and the corresponding weights are learned jointly using a discriminative (supervised) criterion. 
% The parameters to be learned include the low-dimensional representations for various feature parts as well as the weights on these features, i.e. the Tucker approximation. Furthermore, the learning
% Therefore, the proposed approach is learning these projections along with the weights. And instead of finding a Tucker approximation with respect to an unsupervised objective based on the squared error loss function, we are finding one that is the most discriminative (i.e. w.r.t. a supervised criterion). 
We call the model based on this representation the {\emph{Low-Rank Feature Representation with Tucker form}}, or \lrfcmtngram. %, where $n$ indicates the number of lexical parts.

\subsection{CP Form}
% \subsection{Handling Multiple Lexical Parts with CP Approximation}
\label{ssec:lr_ngram}
%In order to handle features with multiple lexical parts ($n$ words), we consider tensor models of higher orders ($n+2$), where each of the first $n$ views corresponds to a word:
%% \SideNoteMY{\raman{Earlier we used $m$ for $n$ here which I think was a typo but  please confirm} No we always use $n$ to make it consistent with $n$-grams}
%\begin{eqnarray}
%% \textstyle \!P(\y| \bx;\! \mathbf{e},\! g,\!\mathbf{\{\w_i\}}, \!\mathbf{\w_f}, \!\mathbf{\w_\ell}) \! \propto\! \exp\Big(\sum_i g \times_\ell  \nonumber \\ 
%  \textstyle \!s({\w}_{1,\cdots,n}, \x, \y;\! \mathbf{e}, g,{\{\bW{i}\}},{\bW{f}},{\bW{\ell}})  = g \times_\ell \nonumber\\ 
% \quad \bh_\y \times_f \mathbf{h}_f \times_{\w_1} \bh_{\w_1} \times_{\w_2} \bh_{\w_2}\cdots\times_{\w_n} \bh_{\w_n}. 
%  \label{eq:fcm_ngram}
%\end{eqnarray}
% \SideNoteMY{Why $\otimes$ here?}

For the Tucker approximation the number of parameters in \eqref{eq:fcm_ngram} scale exponentially with the number of lexical parts.
 For instance, suppose each $\bh^{(i)}_{\w_i}$ has dimensionality $r$, then $\vert g\vert \propto r^n$. To address scalability and further control the complexity of our tensor based model, we approximate the parameter tensor using  CP decomposition as in  \eqref{eq:cp_decomp},
%  i.e. we approximate $\cT = \bW{\l} \otimes \bW{f} \otimes \bW{1} \otimes \cdots \otimes \bW{n},$ as in \eqref{eq:cp_decomp}, 
resulting in the following scoring function:
\begin{align}
%  &\textstyle P(\y | \bx; \mathbf{e}, \mathbf{\{\w_i\}},\mathbf{\w_f}, \mathbf{\w_\ell}) \nonumber \propto\\ 
  \textstyle & \hspace*{-7pt} s(\y, \x, \w_1, \cdots, \w_n; {\bW{\l}}, {\bW{f}}, {\{\bW{i}\}_{i=1}^n})  = \nonumber\\ 
 &  \qquad  \sum_{j=1}^r \left(\bh^{(\l)}_\y \circ \mathbf{\bh^{(f)}_{\x}}   \circ \bh^{(1)}_{\w_1} \circ   \cdots \circ \bh^{(n)}_{\w_n} \right)_j. 
  \label{eq:fcmcp_ngram}
\end{align}
We call this model \emph{Low-Rank Feature Representation with CP form} (\lrfcmcpngram). 

\subsection{Pre-trained Word Embeddings}
\label{ssec:lr_embedding}
One of the computational and statistical bottlenecks in learning these \lrfcmngram~models is the vocabulary size; the number of parameters to learn in each matrix $\bW{i}$ scales linearly with $|V|$ and would require very large sets of labeled training data. To alleviate this problem, we use pre-trained continuous word embeddings~\cite{mikolov2013distributed} as input embeddings rather than the one-hot word encodings. We denote the $m$-dimensional word embeddings by $e_\w$; \SideNoteRA{I used the same notation $e_\w$ as you use later for word2vec embeddings. However, I feel it would be more appropriate to use $e_\w$ for one-hot encoding as in previous subsection and use $\bW{i}$ or $\v$ for word2vec embeddings. Also I would use bold $e$ to be consistent.} so the transformation matrices $\bW{i}$ for the lexical parts are of size $r_i \times m$ where $m \ll |V|.$ 
\removed{
To alleviate this problem, we adopt a two-step approach to get the hidden representation of word
$h_{\w_i}$:
we first transform a word to a $d_{emb}$-dimensional ($d_{emb} \ll |V|$) continuous embedding
with a set of \emph{pre-trained word embedding} %(denoted as original embeddings in the rest of the paper)
 $\be$ from word2vec \cite{mikolov2013distributed}.
Then we use the matrix $\bW{i}$ to get the (task-specific) hidden representation 
$\bh_{\w_i}=\bW{i} e_{\w_i}$, where $e_{\w_i}$ is the pre-trained word embedding vector of $w_i$. 
In this situation the matrix $\bW{i}$ to learn has the size $d_{emb} \times r_i$, 
significantly smaller.}

We note that when sufficiently large labeled data is available, our model allows for fine-tuning the pre-trained word embeddings to improve the expressive strength of the model, as is common with deep network models.

%\subsection{Remarks}
\paragraph{Remarks}
%For unigram lexical features, our models, especially \lrfcmcp,  due to the dimension reduction on each view.
%Such
Our \lrfcm{s} introduce embeddings for non-lexical properties and
labels, making them better suit the common setting in NLP: rich linguistic properties; and large label sets such as open-domain tasks \cite{hoffmann-zhang-weld:2010:ACL}. 
The \lrfcmcp\ better suits $n$-gram features, since when $n$ increases 1, the only new parameters are the corresponding $\bW{i}$.
It is also very efficient during prediction ($O(nr)$), since the cost of transformations can be ignored with the help of look-up tables and pre-computing.

\section{Learning Representations for $n$-gram Lexical Features of Mixed Lengths}
\label{sec:lr_model_ngram}

For features with $n$ lexical parts, we can train an \lrfcmngram\ model to obtain their representations.
However, we often have features of varying $n$ (e.g. both unigrams
($n$=1) and bigrams ($n$=2) as in Figure \ref{fig:feat_template}).
%We will not want to ignore $n$-gram features with some number $n$, so 
%we seek representations for features with arbitrary lengths of $n$ simultaneously. 
We require representations for features with arbitrary different $n$ simultaneously.  
%So far we have assumed that all features rely on $n$-grams of the same length, where the $n$-gram was divided into
%$n$ lexical parts.
%But models often use features with varying numbers of lexical parts,
%such as including features with both unigrams and bigrams.
%We seek representations for features with arbitrary lengths of $n$ simultaneously. 

We propose two solutions. The first is a straightforward solution based on our framework, which handles each $n$ with a $(n+2)$-way tensor. This strategy is commonly used in NLP, e.g. \newcite{taubtabib-goldberg-globerson:2015:NAACL-HLT} have different kernel functions for different order of dependency features. The second is an approximation method which aims to use a single tensor to handle all $n$s.
 
\setlength{\abovedisplayskip}{4pt}
\setlength{\belowdisplayskip}{4pt}
 
%\paragraph{Optimization Objective for an Arbitrary Feature Set}
\paragraph{Multiple Low-Rank Tensors}
Suppose that we can divide the feature set $S(\bx, \y)$ into subsets $S_1(\bx, \y), S_2(\bx, \y), \ldots, S_n(\bx, \y)$ which correspond to features with one lexical part (unigram features), two lexical parts (bigram features)$, \ldots$ and $n$ lexical parts ($n$-gram features), respectively.
To handle these types of features, we modify the training objective as follows:
%\vspace{-0.1in}
\begin{align}
\underset{\cT_1,\cT_2,\cdots,\cT_n}{\textrm{minimize}}{\sum_{(\bx,\y)\in D} \ell(\bx, \y;\cT_1, \cT_2, \ldots,...\cT_n)},
\label{eq:objective_multiple}
\end{align}
where the score of a training instance $(\bx, \y)$ is defined as $s(\bx,\y;\cT) = \sum_{i=1}^n \sum_{\phi \in S_i(\bx,\y)} s(\phi;\cT_i)$.
We use the Tucker form low-rank tensor for $\cT_1$, and the CP form for $\cT_i$ $(\forall i>1)$. We refer to this method
as \lrfcm$_1${\sc -tucker} \& \lrfcm$_2${\sc -cp}.

%（2）特征向量应该在小的子空间中，（3）利用structured sparsity。写in general可以用group sparsity，但是这里我们直接用template来控制sparsity。

\paragraph{Word Clusters}
\label{ssec:lr_clusters}
\removed{
We now propose an alternative solution for feature representations with multiple
lexical parts. Instead of using the CP form tensor approximation in \S\ref{ssec:lr_ngram}, 
we instead replace some lexical parts with discrete word clusters.
%so they can be reduced
%to non-lexical parts.
}
Alternatively, to handle different numbers of lexical parts, we
%An alternative to handling different numbers of lexical parts using multiple low-rank tensors is to 
replace some lexical parts with discrete word clusters. 
%so they can be reduced
%to non-lexical parts.
Let $c(\w)$ denote the word cluster (e.g. from Brown clustering) for word $\w$. For bigram features we have:
%
%\vspace{-0.1in}
\begin{align}
&\hspace*{1pt}  \textstyle s(\y, \x, {\w}_1, \w_2; {\cT})  \nonumber\\ 
 & \textstyle = s(\y, \x \!\wedge\! c(\w_1), \w_2; {\cT}) + s(\y, \x \!\wedge\! c(\w_2), \w_1; {\cT}) \nonumber\\ 
  &\textstyle = {\cT} \times_{\l} \y \times_f \left(\x \wedge c(\w_1) \right) \times_{\w} e_{\w_2} \nonumber\\
  &\textstyle \qquad + {\cT} \times_{\l} \y \times_f \left(\x \wedge c(\w_2) \right) \times_{\w} e_{\w_1} 
  \label{eq:fcmcluster_ngram}
\end{align}
where for each word we have introduced an additional set of non-lexical properties that are conjunctions of word clusters and the original non-lexical properties. % thereby reducing $n$-gram representations to unigram.
This allows us to reduce an $n$-gram feature representation to a unigram representation. 
% In this way we reduce any $n$-gram feature representation learning to unigram feature representations.
%where $c(\w)$ is $\w$'s the word cluster (e.g. from Brown clustering). $\hat{\cT}$ is the Tucker form low-rank tensor as in \eqref{eq:fcm_ngram}.
%For each word, we introduce an additional sets of non-lexical features which are conjunctions of word clusters
%and the original non-lexical features.
The advantage of this method is that it uses a single low-rank tensor to score features with different numbers of lexical parts. This is particularly helpful when we have very limited labeled data. %, since it has fewer overall parameters.
%We will compare this method to \lrfcmcpngram\ in
%the experiments.
We denote this method as \lrfcmcluster, since we use Brown clusters in practice.
In the experiments we use the Tucker form for \lrfcmcluster.

\NoteMRG{Should we also mention the strawman which simply uses
  the max $n_{max}$ of any feature and expands the features with $n <
  n_{max}$ by adding dummy words that always fire?}
\NoteMY{No need for dummy words actually. For example, eq (9) demonstrates the bigram case. Then for the unigram case, we use the same tensor T. And for each word, the non-lexical features just do not contain any Brown clusters. But they are still non-lexical features so we can put them and the \x in eq(9) in the same space }
\NoteMRG{Right. But I guess the setting I'm proposing would not use
  Brown clusters, it would use the actual word types.}

\section{Parameter Estimation}
\label{ssec:train}

The goal of learning is to find a tensor $\cT$ that solves problem~\eqref{eq:objective}. 
Note that this is a non-convex objective, so compared to the convex objective in a traditional log-linear model, we are trading better feature representations with the cost of a harder optimization problem.
While stochastic gradient descent (SGD) is a natural choice for learning representations in large data settings, problem~\eqref{eq:objective} involves rank constraints, which require an expensive proximal operation to enforce the constraints at each iteration of SGD. We seek a more efficient learning algorithm. Note that we fixed the size of each  transformation matrix $\bW{i} \in\R^{r_i\times d_i}$ so that the smaller dimension ($r_i$ $<$ $d_i$) matches the upper bound on the rank. Therefore, the rank constants are always satisfied through a run of SGD and we in essence  have an unconstrained optimization problem. Note that in this way we do not guarantee orthogonality and full-rank of the learned transformation matrices. These properties are assumed in general, but are not necessary according to \cite{kolda2009tensor}.

The gradients are computed via the chain-rule. We use AdaGrad \cite{duchi2011adaptive} and apply L2 regularization on all $\bW{i}$s and $g$, except for the case of $r_i$=$d_i$, where we will start with $\bW{i}=\mathbf{I}$ and regularize with $\|\bW{i}$ - $\mathbf{I} \|_2$. We use early-stopping on a development set.% to alleviate over-fitting.

% We should remark though that this 
%
% adopt a different approach and remove the constraints to simplify the learning objective; 
%The key idea is to pre-set the sizes of all $\bW$s of the model, therefore the upper bound of ranks of
%the learned model can be fixed. Then as long as this upper bound can satisfy the constrains, all
%the possible sets of parameters can be learned will satisfy the constrains.
%the low-rank approximation method of~\cite{socher-EtAl:2012:EMNLP-CoNLL}. 

%\input{Training}

\section{Experimental Settings}
\label{sec:setting}

\begin{table*}[htbp]
\centering
\scriptsize
%\small
\begin{tabular}{|l|c|c|c|c|}
\hline
\multirow{2}{*}{\bf Task} & \multirow{2}{*}{\bf Benchmark} & \multirow{2}{*}{\bf Dataset} & \multicolumn{2}{|c|}{\bf Numbers on Each View} \\
	\cline{4-5}
       & &  & \bf \#Labels ($d_1$) & \bf \#Non-lexical Features ($d_2$)\\
       \hline
       Relation Extraction & \newcite{Yu:2015rt} & ACE 2005 & 32 & 264 \\
%       \hline
%       \hline
      PP-attachment & \newcite{belinkov2014exploring} & WSJ & - & 1,213 / 607 \\
      Preposition Disambiguation & \newcite{Ritter:2014learningsemantics} & \newcite{Ritter:2014learningsemantics} & 6 &  9/3\\
%      Phrase Similarity & - & PPDB & 10,000 & 2,354 \\
       %joint-\lrfcm\ (ST) &  & \textbf{} &   \\
        \hline
\end{tabular}
%\vspace{-.1in}
\caption{\small {Statistics of each task. PP-attachment and preposition disambiguation have both unigram and bigram features. Therefore we list the numbers of non-lexical properties for both types.}}
% The bigram model has half number of features as the unigram model has (excluding the bias feature), because there is no need to indicate whether the role ($c(w)$) of a word is the child (C) or head (H).} }
%\vspace{-.1in}
\label{tab:exp_stats}
\end{table*}

\begin{table*}
\centering
\scriptsize
\begin{tabular}{|l|c|}
\hline
\bf Set & \bf Template\\
\hline
\feat{HeadEmb} & $\{I[i=h_1], I[i=h_2]\}$ (head of $M_1/M_2$) \\
&  $\& \{\phi, t_{h_1},t_{h_2},t_{h_1} \& t_{h_2}\}$\\
\hline
\feat{Context} &  $I[i=h_1/h_2\pm 1]$ (left/right token of $w_{h_1/h_2}$) \\
%&  $I[i=h_2\pm 1]$ (left/right token of $w_{h_2}$) \\
\hline
 \feat{In-between}  & $I[i > h_1] \& I[i < h_2]$$\& \{\phi, t_{h_1},t_{h_2},t_{h_1}\& t_{h_2}\}$\\
 \hline
 \feat{On-path} & $I[w_i \in P]$ $\& \{\phi, t_{h_1},t_{h_2},t_{h_1}\& t_{h_2}\}$ \\
 \hline
 \hline
 \bf Set &{\bf Template} \\
	\hline
	\feat{Bag of Words} & $w$, $p$ $\&$ $w$ ($w$ is $w_m$ or $w_h$)\\
	\feat{Word-Position} & $w_m$, $w_h$, $w_m$ $\&$ $w_h$ \\
	\feat{Preposition}& $p$, $p$ $\&$ $w_{m}$, $p$ $\&$ $w_h$, $p$ $\&$ $w_m$ $\&$ $w_h$ \\
	\hline
\end{tabular}
\quad
\begin{tabular}{|c|c|}
	\hline
	\bf Set &{\bf Template}  \\
	\hline
	Bag of Words & $w$ ($w$ is $w_m$ or $w_h$), $w_m \& w_h$\\
	Distance & Dis$(w_h, w_m)$ $\&$ $\{w_m, w_h, w_m \& w_h\}$\\
	Prep & $w_p$ $\&$ $\{w_m, w_h, w_m \& w_h\}$\\
	POS & $t(w_h)$ $\&$ $\{w_m, w_h, w_m \& w_h\}$\\
	NextPOS & $t(w_{h+1})$ $\&$ $\{w_m, w_h, w_m \& w_h\}$ \\
	{VerbNet} & $P=\{p(w_h)\}$ $\&$ $\{w_m, w_h, w_m \& w_h\}$\\
	& $I[w_p \in P]$ $\&$ $\{w_m, w_h, w_m \& w_h\}$\\
	WordNet & $R_h=\{r(w_h) \}$ $\&$ $\{w_m, w_h, w_m \& w_h\}$\\
	& $R_m=\{r(w_m) \}$ $\&$ $\{w_m, w_h, w_m \& w_h\}$\\
	\hline
	\end{tabular}
%\quad
%\begin{tabular}{|c|c|c|}
%	\hline
%	\bf Set &{\bf Template} \\
%	\hline
%	Bag of Words & $w$, $p$ $\&$ $w$ ($w$ is $w_m$ or $w_h$)\\
%	Words \& positions & $w_m$, $w_h$, $w_m$ $\&$ $w_h$ \\
%	Preposition& $p$, $p$ $\&$ $w_{m}$, $p$ $\&$ $w_h$, $p$ $\&$ $w_m$ $\&$ $w_h$ \\
%	\hline
%	\end{tabular}
\caption{\small{\textbf{Up-left}: Unigram lexical features (only showing non-lexical parts) for \textbf{relation extraction} (from Yu et al. (2014)). We denote the two target
entities as $M_1,M_2$ (with head indices $h_1,h_2$, NE types $t_{h_1},t_{h_2}$), and their dependency path as $P$.}
{\small \textbf{Right}: Uni/bi-gram feature for \textbf{PP-attachment}: 
Each feature is defined on tuple ($w_m$, $w_p$, $w_h$), where $w_p$ is the preposition word,
$w_m$ is the child of the preposition, and $w_h$ is a candidate head of $w_p$.
%$m$ and $h$ are the absolute positions of the two words in a sentence. 
$t(w)$: POS tag of word $w$; $p(w)$: a preposition collocation of verb $w$ from VerbNet; $r(w)$: the root hypernym of word $w$ in WordNet. Dis$(\cdot, \cdot)$: the number of candidate heads between two words. }
{\small \textbf{Down-left}: Uni/bi-gram feature for \textbf{preposition disambiguation} (for each preposition word $p$, its modifier noun $w_m$ and head noun $w_h$). 
%The bag-of-words features ignore the positions of each word.
Since the sentences are different from each other on only $p$, $w_m$ and $w_h$, we ignore the words on the other positions.}
}
\label{tab:feat_template}
%\vspace{-.2in}
\end{table*}

%\begin{table*}
%\parbox{.45\linewidth}{
%\centering
%\begin{tabular}{ccc}
%\hline
%a&b&c\\
%\hline
%\end{tabular}
%\caption{Foo}
%}
%\hfill
%\parbox{.45\linewidth}{
%\centering
%\begin{tabular}{ccc}
%\hline
%d&e&f\\
%\hline
%\end{tabular}
%\caption{Bar}
%}
%\end{table*}

We evaluate \lrfcm\  on three tasks: 
relation extraction, PP attachment and preposition disambiguation
(see Table \ref{tab:exp_stats} for a task summary).
%Additional experiments on phrase similarity can be found in Appendix F.
We include 
%detailed feature templates and the data statistics are included in Appendix B, C, D and E.
detailed feature templates in Table \ref{tab:feat_template}.

PP-attachment and relation extraction are two fundamental NLP tasks,
and we test our models on the largest English data
sets.\SideNoteMRG{It's possible the Czech treebank would offer the
  largest existing PP-attachment task. Playing it safe and switching
  to \emph{largest English data sets}}
The preposition disambiguation task was designed for compositional semantics, which is an important application of deep learning and distributed representations.
On all these tasks, we compare to the state-of-the-art.% to the best of our known.

We use the same word embeddings in \newcite{belinkov2014exploring} on PP-attachment for a fair comparison.
For the other experiments, we use the same 200-$d$ word embeddings in \newcite{Yu:2015rt}.
%All experiments use 200 dimensional embeddings trained on the NYT
%portion of the Gigaword 5.0 corpus \cite{parker2011english}, with the
%default setting of word2vec 
%\cite{mikolov2013distributed}. 

\paragraph{Relation Extraction}
We use the English portion of the ACE 2005 relation extraction dataset \cite{walker_ace_2006}.
%We train on the union of the news domains (Newswire and Broadcast News), and split in half the
%Broadcast Conversation domain for development and test.
%Following \newcite{sun_semi-supervised_2011}, 
%we report results when both gold entity spans
%and types are available during train and test.
Following \newcite{Yu:2015rt}, we use both gold entity spans and types, train the model on 
the news domain and test on the broadcast conversation domain.
%Newswire and Broadcast News and test on the Broadcast Conversation domain.
To highlight the impact of training data size we evaluate with all 43,518 relations (entity mention pairs)
and a reduced training set of the first 10,000 relations. 
%Besides the baseline method of \cite{Mo-Yu:2014qv}, we also compare with 
%traditional feature-based methods \cite{sun_semi-supervised_2011}.
%Dependency parses are from Stanford CoreNLP \cite{manning-EtAl:2014:P14-5}. 
We report precision, recall, and F1.

We compare to two baseline methods:
1) a log-linear model 
with a rich binary feature set from
\newcite{sun_semi-supervised_2011} and \newcite{zhou_exploring_2005} as
described in \newcite{Yu:2015rt} ({\sc Baseline});
2) the embedding model (\fct{}) of \newcite{gormley-yu-dredze:2015:EMNLP}, which uses
rich linguistic features for relation extraction.
We use the same feature templates
and evaluate on fine-grained relations (sub-types, 32 labels) \cite{Yu:2015rt}.
This will evaluate how \lrfcm{} can utilize non-lexical linguistic features.

\SideNoteMRG{I cut the state-of-the-art modifier on log-linear model
  above. Isn't the LRFCM (Yu et al., 2015) the state-of-the-art for
  the 32 label task?}
%Learning rates $\eta$ and the weight of L2 regularizers $\lambda$ for \lrfcm{} were tuned on the dev set.\footnote{\fcm{} does not
%benefit from L2 regularization.} The chosen values
%were $\eta=0.05$ and $\lambda=0.005$, and the third \lrfcm{} model in Table \ref{tab:res_re} had $\lambda=0.05$. We select the rank with a grid search on the dev set from the following values:
%$r_1=\{10,20,d_1\}$, $r_2=\{20,50,d_2\}$ and $r_3=\{50,100,200\}$.
%In this paper when $r_i=d_i$, we will start with $\bW_i=\mathbf{I}$ and
%regularize with L2-norm between the learned $\bW_i$ and $\mathbf{I}$.
%This is actually reducing the complexity of models so it achieve best results on some settings.

% contains
%tens of millions of automatically extracted paraphrase pairs, including words and phrases.
%We extract all paraphrases containing a bigram noun phrase and a noun word from PPDB.

%\SideNoteMY{Other applications: distant supervision, preposition ambiguity and Dependency parsing}
%\SideNoteMY{Put feature templates here (or as Appendix).}

\paragraph{PP-attachment}
We consider the prepositional phrase (PP) attachment task of 
\newcite{belinkov2014exploring},\footnote{\tiny{\url{http://groups.csail.mit.edu/rbg/code/pp}}}
%/pp-code-data.tar.gz}}},
where for each PP the correct head (verbs or nouns) must be selected from content words before the PP (within a 10-word window).
We formulate the task as a ranking problem,
%(See Appendix D for details and baselines).
where we optimize the score of the correct head from a list of candidates with varying sizes.
%Since the system outputs a ranked list over candidates. 
%Since there are less than 10 candidates, 
%we adopt a list-wise training scheme. Given a 
%preposition $p$ and its child $c$, we minimize the loss of selecting
%the correct head $h$ from the candidate list $L$:
%\begin{align}
% -\log \left( { \exp\left \{ s(h,p,c) \right \}} / {\sum_{h'} \exp\left \{ s(h',p,c) \right \}} \right). \nonumber
%\end{align}
%The unlexical part has the form $f(h,p,c)$, making $\bh_f$  depend on $h$. The scoring function 
%\eqref{eq:train-equivalent} becomes:
%\begin{align}
%&s(y|\bx;\cT) = \sum_{i} g \times_y \bh_y \times_f \bh_{\mathbf{f}_{w_i,y}} \times_w \bh_{w_i}\nonumber .
%%&  = \exp\left \{ (g \times_y \bh_y)  \odot \left( \sum_i \bh_{\mathbf{f}_{w_i}} \otimes \bh_{e_{w_i}} \right) \right \}
%\end{align}

PP-attachment suffers from data sparsity because 
of bi-lexical features, which we will model with methods in \S\ref{sec:lr_model_ngram}.
%using \lrfcmngram{} and \lrfcmcluster{}.
Belikov et al. show that rich features -- POS, WordNet and VerbNet -- help this
task. The combination of these features give a large number of non-lexical properties, for which
%(large $m_2$ in Table \ref{tab:models}).
embeddings of non-lexical properties in \lrfcm\ should be useful.

%Since Belikov et al. did not release their dev set, 
We extract a dev set
from section 22 of the PTB following the description in \newcite{belinkov2014exploring}.
%For \lrfcmt\ we tune the same set of hyper-parameters as in the \emph{relation extraction} task,
%except that there is no $r_1$ in this model.
%We compare with all the results reported in their paper.
%For \lrfcmcp, we select $r=\{50,100,200\}$.

\paragraph{Preposition Disambiguation} We consider the preposition disambiguation task proposed by \newcite{Ritter:2014learningsemantics}.
The task is to determine the spatial relationship a preposition indicates based on the two objects connected by the preposition.
For example, ``the apple on the refrigerator'' indicates the ``support by Horizontal Surface'' relation, while ``the apple on the branch''
indicates the ``Support from Above'' relation. 
%This can be formulated as a classification task for noun phrases composed of a noun phrase
%(e.g ``the apple") and a prepositional phrase (e.g. ``on the branch'').
Since the meaning of a preposition depends on the combination of both its head and child word,
we expect conjunctions between these word embeddings to help, i.e. features with two lexical parts.

We include three baselines: 
point-wise addition (SUM)~\cite{mitchell2010composition},
concatenation~\cite{Ritter:2014learningsemantics}, and
%Ritter et al. and
an SVM based on hand-crafted features in Table~\ref{tab:feat_template}.
Ritter et al. show that the first two methods beat other
compositional models.
%For hyper-parameter tuning of \lrfcmt, we use the same setting as in relation extraction,
%but do not choose $r_1$ since the number of labels is small.
%For \lrfcmcp, we select $r=\{50,100,200\}$.

\paragraph{Hyperparameters} are all tuned on the dev set.
The chosen values are learning rate $\eta=0.05$ and the weight of L2
regularizer $\lambda=0.005$ for \lrfcm{}, except for the third \lrfcm{} in Table \ref{tab:res_re} which has $\lambda=0.05$.
%\footnote{\fcm{} does not benefit from L2 regularization.}.
We select the rank of \lrfcmt\ with a grid search from the following values:
$r_1=\{10,20,d_1\}$, $r_2=\{20,50,d_2\}$ and $r_3=\{50,100,200\}$. For \lrfcmcp, we select $r=\{50,100,200\}$.
For the \emph{PP-attachement} task there is no $r_1$ since it uses a ranking model.
For the \emph{Preposition Disambiguation} we do not choose $r_1$ since the number of labels is small.

\section{Results}
\label{sec:exp}

\NoteMRG{The dashed lines in the table are confusing.}
\NoteMRG{Maybe we should remove the horizontal lines from the tables altogether?}

\begin{table*}[htbp]
\centering
\small
\begin{tabular}{|l|c|c|c|c|c|c|c|c|c|c|}
\hline
\multirow{2}{*} & \multicolumn{3}{|c|}{\bf Parameters} & \multicolumn{3}{|c|}{\bf Full Set ($\vert D \vert$=43,518)} & \multicolumn{3}{|c|}{\bf Reduced Set ($\vert D \vert$=10,000)} &\bf Prediction\\
\cline{2-10}
\bf Method & \bf $\mathbf{r_1}$ & \bf $\mathbf{r_2}$ & \bf $\mathbf{r_3}$ & \bf P & \bf R & \bf F1 & \bf P & \bf R & \bf F1&\bf Time (ms) \\
       \hline
       %\newcite{sun_semi-supervised_2011} (ST) & \bf 72.2 & 52.0 &\bf 60.5 & 60.2 & \bf 51.2 & 55.3 &  & & & & & \\
       {\sc Baseline} & \bf - & - &\bf - & 60.2 & \bf 51.2 & 55.3 & - & - & - & -\\
       \fct\  & 32/N & 264/N & 200/N & 62.9 &49.6 & 55.4 & \bf 61.6 & 37.1 & 46.3 & 2,242\\
       \hline
       \lrfcm$_1${\sc -tucker}\  & 32/N & 20/Y & 200/Y & 62.1 & \bf 52.7 & \bf 57.0 & 51.5 & 40.8 & 45.5 & 3,076\\
       \lrfcm$_1${\sc -tucker}\  & 32/N & 20/Y & 200/N &\bf 63.5 & 51.1 &  56.6 &  52.8 & 40.1 & 45.6 &  2,972\\
%       \hdashline
       \lrfcm$_1${\sc -tucker}\  & 20/Y & 20/Y & 200/Y & 62.4 & 51.0 & 56.1 & 52.1 & 41.2 &  46.0 & 2,538\\
       \lrfcm$_1${\sc -tucker}\  & 32/Y & 20/Y & 50/Y &57.4 & 52.4 & 54.8 & 49.7 & \bf 46.1 & 47.8 & 1,198\\
%       \hline
       \hdashline
       \lrfcm$_1${\sc -cp}\ & \multicolumn{3}{|c|} {200/Y} &61.3 & 50.7 & 55.5 & 58.3 & 41.6 & \bf 48.6 & \bf 502 \\
       %joint-\lrfcm\ (ST) &  & \textbf{} &   \\
        \hline
\end{tabular}
%\begin{tabular}{|l|c|c|c|c|c|c|c|c|c|c|}
%\hline
%\multirow{2}{*} & \multicolumn{3}{|c|}{\bf Parameters} & \multicolumn{3}{|c|}{\bf Full Set ($\vert D \vert$=43,518)} & \multicolumn{3}{|c|}{\bf Reduced Set ($\vert D \vert$=10,000)} &\bf Prediction\\
%\cline{2-10}
%\bf Method & \bf $\mathbf{r_1}$ & \bf $\mathbf{r_2}$ & \bf $\mathbf{r_3}$ & \bf P & \bf R & \bf F1 & \bf P & \bf R & \bf F1&\bf Time (ms) \\
%       \hline
%       %\newcite{sun_semi-supervised_2011} (ST) & \bf 72.2 & 52.0 &\bf 60.5 & 60.2 & \bf 51.2 & 55.3 &  & & & & & \\
%       {\sc Baseline} & \bf - & - &\bf - & 60.2 & \bf 51.2 & 55.3 & - & - & - & -\\
%       \fct\  & 200/N & 264/N & 32/N & 62.9 &49.6 & 55.4 & \bf 61.59 & 37.09 & 46.3 & 2242\\
%       \hline
%       \lrfcm$_1${\sc -tucker}\  & 200/N & 20/Y & 32/Y & 62.1 & \bf 52.7 & \bf 57.0 & 51.52 & 40.76 & 45.51 & 3,076\\
%       \lrfcm$_1${\sc -tucker}\  & 200/N & 20/Y & 32/N &\bf 63.5 & 51.1 &  56.6 &  52.79 & 40.06 & 45.55 &  2,972\\
%       \hdashline
%       \lrfcm$_1${\sc -tucker}\  & 200/N & 20/Y & 20/Y & 62.4 & 51.0 & 56.1 & 52.05 & 41.18 &  45.98 & 2,538\\
%       \lrfcm$_1${\sc -tucker}\  & 50/Y & 20/Y & 32/Y &57.41 & 52.39 & 54.79 & 49.70 & \bf 46.12 & 47.84 & 1,198\\
%       \hline
%       \lrfcm$_1${\sc -cp}\ & \multicolumn{3}{|c|} {200/Y} &61.3 & 50.7 & 55.5 & 58.3 & 41.61 & \bf 48.56 & \bf 502 \\
%       %joint-\lrfcm\ (ST) &  & \textbf{} &   \\
%        \hline
%\end{tabular}
%\vspace{-.1in}
\caption{\small Results on test for relation extraction. Y(es)/N(o) indicates whether embeddings are updated during training.
%Lines 3 \& 4 are the two best sets of hyper parameters selected under the the full setting. Line 5 \& 6 are the best hyper 
%parameters selected under the low-resource setting.
} 
\label{tab:res_re}
%\vspace{-.1in}
%\vspace{-.2cm}
\end{table*}

\begin{table*}[htbp]
\centering
\small
\begin{tabular}{|l|c|c|}
\hline
\bf System & \bf Resources Used & \bf Acc\\
\hline
%Closest & - & distance & 81.7 \\
%\hline
\multirow{1}{*}{SVM} \cite{belinkov2014exploring} & distance, word, embedding, clusters, POS, WordNet, VerbNet &\multirow{1}{*}{86.0} \\
%\multirow{1}{*}{SVM} & smoothed full features & distance, clusters, POS, WordNet, VerbNet &\multirow{1}{*}{} \\
%\hline
%{HPCD (basic)} && 85.4\\
HPCD \cite{belinkov2014exploring} & distance, embedding, POS, WordNet, VerbNet&{88.7}\\
%\hline
%\multirow{1}{*}{\lrfcm$_1${\sc-tucker}} & smoothed unigram features & distance,embedding, POS,next POS,VerbNet  & 89.8 \\ %87.85 \\
%& + POS,next POS,VerbNet  & 90.1 \\ %90.06\\
%\hline
%\multirow{1}{*}{\lrfcm$_2${\sc-cp}}& smoothed bigram features & distance,embedding, POS,next POS,VerbNet  &  90.1\\
%& + POS,WordNet,VerbNet & 90.1\\ %90.12\\
\hline
{\lrfcm$_1${\sc-tucker}} \& {\lrfcm$_2${\sc-cp}}& distance, embedding, POS, WordNet, VerbNet & \multirow{1}{*}{\bf 90.3}\\ % \multirow{2}{*}{\bf 90.52}\\
\hdashline
{\lrfcmcluster}& distance, embedding, clusters, POS, WordNet, VerbNet & \multirow{1}{*}{89.6}\\ 
\hline
\hline
%Malt & - & \multirow{4}{*}{dependency parser} & 79.7 \\
%MST & - & & 86.8\\
%Turbo &  \multirow{2}{*}{dependency parser} & 88.3\\
RBG \cite{lei-EtAl:2014:P14-1}& dependency parser & 88.4\\
%\hline
%RNN & re-ranker, embedding &85.1 \\
Charniak-RS \cite{mcclosky2006effective} & dependency parser + re-ranker & 88.6\\
%\hline
%\hline
RBG + HPCD (combined model) & dependency parser + distance, embedding, POS, WordNet, VerbNet & \multirow{1}{*}{90.1}\\
\hline
\end{tabular}
%\vspace{-.1in}
\caption{\small PP-attachment test accuracy. The baseline results are from Belinkov et al. (2014).} 
\label{tab:res_ppa}
%\vspace{-.15in}
\end{table*}

%\vspace{-.05in}
\paragraph{Relation Extraction}
All  \lrfcmt\ models improve over {\sc Baseline} and \fct{} (Table \ref{tab:res_re}), making these the best reported
numbers for this task.
However, \lrfcmcp{} does not work as well on the features with only
one lexical part. The Tucker-form does a better job of capturing
interactions between different views.
In the limited training setting, we find that \lrfcmcp{} does best.
%Currently there are also some attempts of applying CP approximation
%(yet not any comparisons between different low-rank approximation methods) 
%to combine features
%from different views \cite{cao-khudanpur:2014:P14-1,lei-EtAl:2014:P14-1}.
%Although we have a different objective here, our results still give inspiration that
%tucker method on their tasks may also improve the results.

Additionally, the primary advantage of the CP
approximation is its reduction in the number 
of model parameters and running time. 
We report each model's running time for a single pass on the development set.
The \lrfcmcp{} is by far the fastest.
The first three \lrfcmt{} models are slightly slower than \fcm{}, 
because they work on dense non-lexical property embeddings while \fcm{} benefits from
sparse vectors.
\paragraph{PP-attachment}
\label{ssec:exp_ppa}
\tabref{tab:res_ppa} shows that \lrfcm{} (89.6 and 90.3) improves over the previous
best standalone system HPCD (88.7) by a large margin, with exactly the
same resources. 
\newcite{belinkov2014exploring} also reported results of parsers and parser re-rankers,
which can access to additional resources (complete
parses for training and complete sentences as inputs) so it is unfair to compare them with the
standalone systems like HPCD and our \lrfcm{}. Nonetheless
{\lrfcm$_1${\sc-tucker}} \& {\lrfcm$_2${\sc-cp}} (90.3) still outperforms 
the state-of-the-art parser RBG (88.4), re-ranker Charniak-RS (88.6),
and the combination
of the state-of-the-art parser and compositional model RBG + HPCD
(90.1). Thus, even with fewer resources, \lrfcm{} becomes the new best system.% on this task.

% MRG: COMMENTED OUT PREVIOUS TWO PARAGRAPHS BELOW TO FOCUS ON PARSER VS.
% STANDALONE DIFFERENCES.
%
% \tabref{tab:res_ppa} shows that \lrfcm{} improves over the previous best single system HPCD (88.7) by a large margin, with exactly the same resource. 
% \lrfcm{} also improves over the SVM based on hand-engineered features (86.0),
% as well as the state-of-the-art parser RBG (88.4) and re-ranker Charniak-RS (88.6),
% becoming the new best single system on this task.
%
% Finally the two \lrfcm{} single systems are even on par with the combination
% of the state-of-the-art parser and compositional model,
% while {\lrfcm$_1${\sc-tucker}} \& {\lrfcm$_2${\sc-cp}} outperforms all the results.
% \NoteMRG{We never define 'single systems' vs. 'combined model' in
%   great detail, so this might be a point of confusion.}

Not shown in the table: we also tried \lrfcm$_1${\sc-tucker} \&
\lrfcm$_2${\sc-cp} with \emph{postag features only} (89.7), and
with grand-head-modifier conjunctions removed (89.3) .
%Comparing to RBG (88.4) \cite{lei-EtAl:2014:P14-1}, our model also relies on low-rank tensors, does not use more resources, but performs better.
Note that compared to \lrfcm, RBG benefits from binary features, which also exploit grand-head-modifier structures.
Yet the above reduced models still work better than RBG (88.4) without
using additional resources.\footnote{Still this is not a fair comparison since we have different training objectives. Using RBG's factorization and training with our objective will give a fair comparison and we leave it to future work.} Moreover, the results of \lrfcm{} can still be potentially improved by combining with binary features.
The above results show the advantage of our factorization method, which allows for utilizing pre-trained word embeddings, and thus can benefit from
semi-supervised learning.
\paragraph{Preposition Disambiguation}

\lrfcm{} improves (Table \ref{tab:res_pp_amb}) over the best methods 
(SUM and Concatenation) in \newcite{Ritter:2014learningsemantics} as well as the SVM
based on the original lexical features (85.1).
In this task \lrfcmcluster\ better represents the
unigram and bigram lexical features, compared to the usage of two low-rank tensors (\lrfcm$_1${\sc -tucker} \& \lrfcm$_2${\sc -cp}). This may be because \lrfcmcluster\ has fewer parameters, which is better for smaller training sets.

We also include a control setting (\lrfcmcluster\ - Control), which has a full rank parameter tensor with
the same inputs on each view as \lrfcmcluster, but represented as one hot vectors
without transforming to the hidden representations $\bh$s.
This is equivalent to an SVM with the compound cluster features as in \newcite{koo_simple_2008}. It performs much worse than \lrfcmcluster, showing the advantage of using word embeddings and low-rank tensors. 
%Note that the method is also worse than the SVM model based on 
%the original feature templates, showing that the usage of word embeddings can better recover 
%the conjunction information between two words.

\begin{table}[t]
\centering
\small
\begin{tabular}{|l|c|c|c|}
\hline
\multirow{1}{*}{ \bf Method} &  \multirow{1}{*}{\bf Accuracy}\\
\hline
SVM - Lexical Features & 85.09\\
%SVM - Brown Features & 75.64\\
%SVM - Control Setting & \\
%\hline
SUM &80.55\\
Concatenation  &  86.73\\
\hline
\lrfcm$_1${\sc -tucker} \& \lrfcm$_2${\sc -cp} & 87.82\\
%\lrfcm$_1${\sc -tucker} & position, prep, cluster, WordNet & \\
%\lrfcm$_2${\sc -tucker} & position, prep & 83.33\\
%\lrfcm$_2${\sc -tucker} & position, prep, cluster & \\
%\lrfcm$_2${\sc -tucker} & position, prep, cluster, WordNet & \\
%\lrfcm$_2${\sc -cp} & position, prep & \\
%\lrfcm$_2${\sc -cp} & position, prep, cluster & \\
%\lrfcm$_2${\sc -cp} & position, prep, cluster, WordNet & \\
\hdashline
\lrfcmcluster  & \bf 88.18\\
\lrfcmcluster\ - Control & 84.18\\
%\hline
%Combined \lrfcm& \bf 89.09\\
\hline
\end{tabular}
%\vspace{-.1in}
\caption{\small Accuracy for spatial classification of PPs.} 
\label{tab:res_pp_amb}
%\vspace{-.2in}
\end{table}

\paragraph{Summary} 
For unigram lexical features, \lrfcmtngram{} achieves better results than
\lrfcmcpngram{}. However, in settings with fewer training examples, features with more lexical parts ($n$-grams), or when
faster predictions are advantageous, \lrfcmcpngram{} does best
as it has fewer parameters to estimate.
For $n$-grams of variable length, \lrfcm$_1${\sc -tucker} \& \lrfcm$_2${\sc -cp} 
does best. In settings with fewer training examples, 
\lrfcmcluster{} does best as it has only one parameter tensor to estimate.

\section{Related Work}
\label{sec:related}

\newcommand{\paragraphAndSentence}[1]{\vspace{.6em} \noindent {\bf #1}}

\paragraph{Dimensionality Reduction for Complex Features} is a
standard technique to address high-dimensional features,
including PCA, alternating structural optimization~\cite{ando2005framework}, denoising autoencoders~\cite{vincent2008extracting}, and feature embeddings~\cite{yang2014unsupervised}. 
These methods treat features as atomic elements and ignore the inner structure of features,
so they learn separate embedding for each feature without shared parameters.
%In other words, the feature representation model will not have shared parameters between features. 
As a result, they still suffer from large parameter spaces when the feature space is very huge.\footnote{For example, a state-of-the-art dependency parser \cite{zhang-mcdonald:2014:P14-2} extracts about 10 million features; in this case, learning 100-dimensional feature embeddings
% as in~\newcite{yang2014unsupervised} 
involves estimating approximately a billion parameters.}
%
%When applied to lexical features for various NLP tasks, these methods suffer from very large parameter spaces. For example, a state-of-the-art dependency parser~\cite{zhang-mcdonald:2014:P14-2} extracts more than 10 million features; in this setting, learning 100-dimensional feature embeddings as in~\cite{yang2014unsupervised} involves estimating approximately a billion parameters, which may be computationally challenging. 

Another line of research studies the inner structures of lexical features:
e.g. \newcite{koo_simple_2008}, \newcite{turian2010word}, \newcite{sun_semi-supervised_2011}, \newcite{nguyen_employing_2014}, \newcite{roth_composition_2014}, and \newcite{hermann-EtAl:2014:P14-1} used pre-trained word embeddings to replace the lexical parts of features~;
%e.g. \newcite{miller_name_2004,koo_simple_2008,turian2010word,sun_semi-supervised_2011,nguyen_employing_2014,roth_composition_2014,hermann-EtAl:2014:P14-1} used pre-trained word embeddings to replace the lexical parts of features~;
%these can therefore be seen as special cases of our model that only work on (the single) lexical parts of features. 
\newcite{srikumar2014learning},~\newcite{gormley-yu-dredze:2015:EMNLP} and~\newcite{Yu:2015rt}
propose splitting lexical features into different parts and employing tensors to perform classification.
%compose representations of these parts. 
The above can therefore be seen as special cases of our model that only embed a certain part (view) of the complex features. This restriction also makes their model parameters form a full rank tensor, resulting in data sparsity and high computational costs when the tensors are large.

\paragraphAndSentence{Composition Models (Deep Learning)}
build representations for
structures based on their component word embeddings
\cite{collobert2011natural,bordes2012semantic,socher-EtAl:2012:EMNLP-CoNLL,socher-EtAl:2013:EMNLP}.
%This way of using word embedding has been shown effective for boosting the performance of many NLP tasks, including syntax
%\cite{collobert2011natural},
%semantics
%\cite{socher-EtAl:2012:EMNLP-CoNLL,socher-EtAl:2013:EMNLP} and question answering \cite{bordes2014question}.
%
%These compositional models aim to build representation of a structure from its sub-structure representations, where word embeddings (can be fine-tuned) serves as ``basic building blocks''.
When using only word embeddings, these models achieved successes on several NLP tasks,
but
%These models usually use only word embeddings. Therefore, despite their successes on several NLP tasks, they 
sometimes fail to learn useful syntactic or semantic patterns beyond the strength of combinations of word embeddings, such as the dependency relation in Figure~\ref{fig:feat_template}(a). 
To tackle this problem,
some work designed their model structures according to a specific kind of linguistic patterns,
e.g. dependency paths
\cite{ma-EtAl:2015:ACL-IJCNLP,liu-EtAl:2015:ACL-IJCNLP},
while a recent trend enhances compositional models
with linguistic features.
% on compositional functions.
% of developing compositional models is to 
%do ``minimal'' feature engineering on compositional functions
%recent trend is to do feature engineering on compositional functions.
%according to their usages in a specific task.
%Therefore the composition of two sub-structures in different 
%situations will be treated differently.
%
For example,
\newcite{belinkov2014exploring} concatenate embeddings with linguistic features 
before feeding them to a neural network;
\newcite{socher2013parsing} and \newcite{hermann2013role} enhanced Recursive Neural Networks by refining
the transformation matrices with linguistic features (e.g. phrase types).
%TODO: dependency parse
%\newcite{zeng-EtAl:2014:Coling} use concatenations of embeddings as features
%in a CNN model, according to their positions relative to the
%target entity mentions.
%\newcite{hermann-EtAl:2014:P14-1}
%gave different weights to word embeddings with different syntactic roles.
%RNN(\cite{socher2013parsing,hermann2013role}). 
%
%Those models reported state-of-the-art performance on the applications 
%on which their papers focused.
These models are similar to ours in the sense of learning representations
based on linguistic features and embeddings.
%However, they can only use limited information 
%(usually one property, such as phrase types in \newcite{socher2013parsing}), 
%limiting the expressiveness of the models.

%At the same time, these models are still easier to get over-fitting
%since they rely on complex deep network structures. 
%On the other hand there may be other different useful properties of a sub-structure which can contribute to the performances of the task.

%As a result, compared to this work, these models (``deep networks'') are still more complex  and easier to get over-fitting,
%due to their ``deep network'' nature (easier to over-fit)
%, while are less expressive than our proposed models,
%as we take advantage of more powerful features with simpler model structures.

\paragraphAndSentence{Low-rank Tensor Models for NLP} aim to handle 
the conjunction among different views of features 
\cite{cao-khudanpur:2014:P14-1,lei-EtAl:2014:P14-1,chen-manning:2014:EMNLP2014}.
\newcite{yu2015learning} proposed a model to compose phrase embeddings from words, which has
an equivalent form of our CP-based method under certain restrictions. 
%The idea of using a tensor to handle 
%the conjunction among different views of features is similar to
%\newcite{cao-khudanpur:2014:P14-1}, \newcite{lei-EtAl:2014:P14-1} and \newcite{chen-manning:2014:EMNLP2014}.
%\footnote{
%Another work related to this paper is \newcite{chen-manning:2014:EMNLP2014}, which embeds both words and some special types of features (POS and dependency labels) and then use a cube activate function to force the embeddings interact with each other.
%%, in order to simulate the traditional features of conjunctions between words and unlexical features. The cube activate function 
%This method can be treated as a special case of 3-order tensor. However,
%it cannot model arbitrary types of features, and are not capable to accurately capture interaction information.},
Our work applies a similar idea to exploiting the inner structure of complex features,
% as non-atomic structures, and to adopt this idea 
and can handle $n$-gram features with different $n$s.
%Our general way of feature decomposition makes our methods
%for compositions of feature embeddings.
%easier to adapt to new tasks.
Our factorization (\S\ref{lab:factor}) is general and easy to adapt to new tasks.
More importantly, it makes the model benefit from pre-trained word embeddings as shown by the PP-attachment results.
%More importantly, it allows for utilizing pre-trained word embeddings, and thus can benefit from
%semi-supervised learning.
%As a result our model can better generalize to unseen data, can be easier to adapt to new tasks,
%to handle features with both single or multiple lexical parts.
%By Mo: about Tucker
%Tucker-form tensor approximation was applied to semantic composition by 
%\newcite{vandecruys-poibeau-korhonen:2013:NAACL-HLT}. 
%We are the first to introduce it to learning feature representations,
%which can better represent features with one lexical part, and can better handle features
%with multiple lexical parts when labeled data is limited.

%\vspace{-.3cm}
\section{Conclusion}

\label{sec:conclusion}
We have presented \lrfcm, a feature representation model that
exploits the inner structure of complex lexical features and 
applies a low-rank tensor to efficiently score features with this representation.
%We have demonstrated that 
\lrfcm\ attains the state-of-the-art on several tasks, including relation extraction, PP-attachment,
and preposition disambiguation. 
We make our implementation available for general use.\footnote{https://github.com/Gorov/LowRankFCM}

%While \lrfcm\  can better utilize small training sets, but it still has many parameters,
%%Although we show that the 
%%the current model still has a large number of parameters to estimate,
%which may limit performance with limited  labeled resources.
%In the future, we will investigate semi-supervised learning of \lrfcm,
%making more parameters (besides word embeddings)
%benefit from unsupervised pre-training.
%% not only the word embeddings, 
%%but also the other parameters of the low-rank approximations.
%%(excluding label embeddings which can only be trained when labeled data are available).

\subsubsection*{Acknowledgements}
\noindent A major portion of this work was done when MY was visiting MD and RA at JHU. This research was supported in part by NSF grant IIS-1546482.

\bibliography{naaclhlt2016.bib}
\bibliographystyle{naaclhlt2016}

\clearpage
%\newpage
%\input{Appendix}

\end{document}